\documentclass[journal,twoside]{ieeecolor}
\usepackage{cite} 
\usepackage{amsmath,amssymb,amsfonts}
\usepackage{graphicx}
\usepackage{textcomp}
\usepackage{wrapfig}
\usepackage{mathtools}
\usepackage{algorithm}
\usepackage{algpseudocode}
\usepackage{epsfig,makeidx,color}
\usepackage{amsmath,amssymb,bbm}
\usepackage{cite,graphicx,lipsum}
\usepackage{enumerate}
\usepackage[switch,pagewise]{lineno}
\usepackage{hyperref}

\usepackage{listings}

\definecolor{codegreen}{rgb}{0,0.6,0}
\definecolor{codegray}{rgb}{0.05,0.05,0.05}
\definecolor{codepurple}{rgb}{0.58,0,0.082}
\definecolor{backcolour}{rgb}{1.,1.,1.}
\definecolor{subsectioncolor}{rgb}{0,0,0}

\lstdefinestyle{mystyle}{
  backgroundcolor=\color{backcolour}, commentstyle=\color{codegreen},
  keywordstyle=\color{magenta},
  numberstyle=\tiny\color{codegray},
  stringstyle=\color{codepurple},
  basicstyle=\fontfamily{pcr}\footnotesize,
  breakatwhitespace=false,         
  breaklines=true,                 
  captionpos=b,                    
  keepspaces=true,                 
  numbers=left,                    
  numbersep=5pt,                  
  showspaces=false,
  showstringspaces=false,
  showtabs=false,                  
  tabsize=2
}

\hypersetup{
        colorlinks = true,
        citecolor=blue,
        linkcolor=blue,
        urlcolor=blue
}
\def\be{ \begin{equation} }
\def\ee{ \end{equation} }
\def\bea{ \begin{eqnarray} }
\def\eea{ \end{eqnarray} }

\def\b0{{\bf 0}}

\definecolor{abstractbg}{rgb}{0.89804,0.94510,0.83137}
\begin{document}

\title{Deakin RF-Sensing: Experiments on Correlated Knowledge Distillation for Monitoring Human Postures with Radios}

\author{Shiva Raj Pokhrel, Jonathan Kua, Deol Satish, Philip Williams, \\ Arkady Zaslavsky,  Seng W. Loke and Jinho Choi \\
\thanks{The authors are with
the IoT Research Lab, School of Information Technology,
Deakin University, Geelong, VIC 3220, Australia.
(e-mail: shiva.pokhrel@deakin.edu.au).
}
}

%\IEEEtitleabstractindextext{%
%\fcolorbox{abstractbg}{abstractbg}{%
%\begin{minipage}{\textwidth}%
%\begin{wrapfigure}[12]{r}{3in}%
%\includegraphics[width=3in]{jsenga.png}%
%\end{wrapfigure}%

\IEEEtitleabstractindextext{%
{%
\begin{minipage}{\textwidth}%
%\begin{wrapfigure}[12]{r}{3in}%
%\includegraphics[width=3in]{jsenga.png}%
%\end{wrapfigure}%
\begin{abstract}\color{black}
In this paper, we study the feasibility of a novel idea by coupling radio frequency (RF) sensing technology with Correlated Knowledge Distillation (CKD) theory towards designing lightweight, near real-time and precise human pose monitoring systems through an in-house experimental testbed development at Deakin University. The datasets collected from the developed testbed are fed to the CKD framework, which transfers and fuses pose knowledge from a robust ``Teacher'' model to a parameterized ``Student'' model and can be a promising technique for obtaining accurate yet lightweight pose estimates. 
As a result, it becomes possible to identify the current pose of a human body (e.g., an elderly individual in a care home) solely using RF signals (e.g., WiFi signals in a home) without relying on visual signals or video information, meaning that this approach effectively addresses privacy concerns by allowing pose identification without compromising personal privacy.
To ensure its efficacy, we implement CKD for distilling logits in our integrated Software Defined Radio (SDR)-based experimental setup and investigate the RF-visual signal correlation. Our CKD-RF sensing technique is characterized by two modes -- a camera-fed Teacher Class Network (e.g., images, videos) with an SDR-fed Student Class Network (e.g., RF signals). Specifically, our CKD model trains a dual multibranch teacher-student network by distilling and fusing knowledge bases. The resulting CKD models are then subsequently used to identify the multimodal correlation and teach the Student branch in reverse. Instead of simply aggregating their learnings, CKD training comprises multiple parallel transformations with the two domains, i.e. visual images and RF signals. Once trained, our CKD model efficiently preserves privacy and utilizes the multimodal correlated logits from the two different neural networks to estimate poses solely using RF signals.
%without using visual signals/video frames (by using only the RF signals).
\end{abstract}

\begin{IEEEkeywords}\color{black}
Radio frequency; Motion detection; RF signals; Sensor phenomena and characterization; Feature extraction; Wireless sensor networks; Machine learning; RF Sensing; Knowledge Distillation; Correlation; Fusion; Software Defined Radio (SDR); RF-based Vision; Pose Monitoring; Body Radio Reflections; RF Localization 
\end{IEEEkeywords}
\end{minipage}}}

\maketitle

\section{Introduction}

The advent of the Internet of Things (IoT), Artificial Intelligence (AI)/Machine Learning (ML), and computer vision have significantly advanced the field of human pose estimation and have resulted in many new application domains, such as near real-time fall detection in patient/aged care homes~\cite{esteva2021deep}, activity sensing~\cite{taylor2022implementation} and sign language recognition~\cite{9187644}. Most state-of-the-art techniques rely on cameras and, more recently, AI-driven computer vision technologies to detect, identify, and react to human pose changes in public/private spaces. These systems can be reliable and provide accurate human pose estimates, but require expensive equipment and incur high computational costs. In addition, there are data privacy concerns when deploying high-definition visual spectrum cameras where the activity of human subjects is being monitored.

\begin{figure}[t]
    \centering
    \includegraphics[width=8cm]{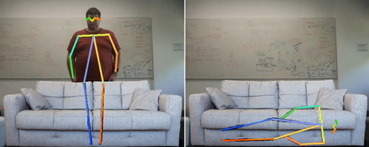}
    \caption{Motivating example of RF-based pose estimation and tracking through occlusions. Human subjects could be hidden from plain view, as RF signals can penetrate objects.}
    \label{fig:motivating}
\end{figure}
The use of Radio Frequency (RF)-sensing technologies for human pose estimation over the past decade has gained traction and has resulted in significant benefits across a wide range of settings (see~\cite{yue2020bodycompass} and references). For example, people in assisted living facilities can benefit from remote monitoring to ensure their health and safety while maintaining their independence, but privacy is invaded by the use of cameras. RF-sensing could not only monitor diseases of the elderly and disabled but also has great potential for military applications (e.g., detecting sitting, standing, and lying face up or down) and emergency response such as fires and flood, including others.%,%10.1145/2486001.2486039,
%zhao2019through}.

Some relevant works directly apply \cite{depatla2018crowd, hanif2018wispy, cheng2019walls, sun2021through, wang2019joint,abdelnasser2018ubiquitous} ML methods for RF-sensing, which have been reported to be strongly biased due to background RF noises and fail to learn helpful information for pose estimates~\cite{taylor2022implementation,li2022unsupervised}. In a different context, \cite{9187644} uses ML to demonstrate the recognition of sign language with RF sensing. However, researchers in this field have poorly studied the potential of multimodal learning with knowledge fusion / distillation to improve RF detection, which is the primary focus of this article. 

Figure~\ref{fig:motivating} illustrates a motivating example of using RF signals to detect and track human poses (the subject could be hidden from plain view, as RF signals can penetrate objects). We aim to develop a system to accurately detect and predict human activity without compromising privacy (movement tracking through occlusions) and alleviating the impacts of RF noise. With computational and wireless network resources becoming increasingly cheaper and available, we design an experimental testbed and propose a new technique by coupling RF sensing with knowledge fusion and distillation. 

The proposed \textbf{Deakin RF-Sensing} is based on the correlated knowledge distillation (CKD) theory, which aims to drive lightweight, near real-time, for precise human pose estimates. Our CKD-based technique uses ``Teacher'' and ``Student'' deep network models to train and implement two modes -- a camera-fed Teacher Class Network (e.g., images, videos) with an RF-fed Student Class Network (e.g., RF signals). Our CKD model trains a dual multibranch teacher-student network and acquires a joint CKD model~\cite{hinton2015distilling,li2021online}, and slightly differently in the decoupled version~\cite{zhao2022decoupled}. The resulting CKD models are subsequently used to identify the multimodal correlation and teach the Student branch in reverse~\cite{peng2019correlation,li2020local}. Instead of simply aggregating their learnings, CKD training always seeks to comprise multiple parallel transformations with the two domains, i.e., visual images and RF signals, as feature mapping obtained by feature fusion seldom serves as a good Teacher to guide their training~\cite{li2023knowledge}. Once trained, using radio signals generated from a Software Defined Radio (SDR), our CKD model can efficiently preserve privacy and utilize the multimodal correlated logits from the Teacher to the Student network without using visual signals/video frames.

 \subsection{Main Novelties and  Contributions}
% {\textcolor{red}{Paragraph 3 Updated:}
\color{black}
We have explored and exploited the software-defined radio (SDR) mechanism to realize a hybrid RF and visual camera architecture for recognizing human actions non-intrusively. To this end, we established an in-house custom experimental testbed to investigate and research the performance of our suggested strategies. We present a novel CKD-based methodology to find and infer the complex relationships between RF and visual camera signals. The RF sensing can be driven separately from the camera's vision system once a CKD model is trained on the data and knowledge and insights about the correlation between the two signals are captured.

We used two Omni-directional VERT2450 antennas and a USRP N200 with a UBX 40 USRP Daughterboard. Each of the two antennas handles reception, while the other handles transmission. We utilized the SDR data for our estimation of posture annotations. Our CKD model is trained (supervised learning, applying proper pose annotations) on the acquired data (approximate stick figures). Combining an RGB camera with Alphapose enables such an accomplishment. Our testing results demonstrate the efficacy of the proposed integrated CKD-RF sensing framework.

Our main contributions to this paper are explained below.
\begin{itemize}
    \item We develop a purpose-built experimental testbed framework that implements a joint SDR and visual camera-based system for near real-time human pose detection and analytics.
    \end{itemize}
    
    \subsubsection{Purpose-built Experimental Testbed} Our testbed network faces unique difficulties in its design and training compared to visual methods of posture examination such as Alphapose~\cite{fang2017rmpe}. In our case, especially with RF signals, there is a lack of labeled data--one cannot realistically add keypoint annotations to RF broadcasts~\cite{zhao2018through}. Therefore, we employ correlated monitoring to solve this issue, simply by pairing the camera with our RF-sensing, thus synchronizing the RF and Video streams during training. In training, we utilize RF stream supervision based on poses extracted from the video frames. Once trained, the system takes the RF as its only input and outputs the results. The end result of the proposed approach is that it can estimate a person's position only from RF signals, without any help from the camera. Moreover, the RF-based framework learns to quantify poses even when humans are entirely obscured, which is an exciting capability, and it achieves this capability despite never being exposed to such situations throughout its training.
    
    \begin{itemize}
    \item We design and propose a novel Correlated Knowledge Distillation (CKD) theory to rapidly track the changes in the physical environment by exploiting the multimodal correlation with fusion between RF and visual signals.
     \end{itemize}
     \subsubsection{Major rethink on KD.} We dig into the details and process of how information gets distilled in the classic~\cite{hinton2015distilling} and decoupled~\cite{zhao2022decoupled} KD. Using ideas from~\cite{zhao2022decoupled}, we recast the KD loss as a weighted sum of class-independent and class-dependent components. In particular, in the proposed CKD, the KD classification prediction problem is divided into two levels: (i) a binary prediction for the camera-based Teacher class and all SDR-based Student classes (RF-sensing), and (ii) a multicategory prediction for each Student class. We use mathematics as a tool to reason the complexity of the developed CKD theory in Sec.~\ref{sec:ckdtheory}. Furthermore, we investigated the impact of each element of the CKD framework and exposed the shortcomings of traditional KD. Using these results as inspiration, we present the new insights of CKD that perform very well in the context of the \textbf{Deakin RF-Sensing} proposed in this paper.
     
    \begin{itemize}
    \item We implement a tractable CKD model for private identification and predictions of human poses without feeding camera images.
\end{itemize}
\subsubsection{Privacy compliance.} The ability of our framework to monitor the whereabouts of people without invading their privacy is a major selling point of this research project. All that can be done to identify people using RF-sensing is to gather their silhouettes, as no information about their faces or other attributes can be transmitted. Moving forward, we are creating a `\textbf{Deakin RF-Sensing spinoff}' product (a commercial application that is secure, efficient, and up to date with the current privacy norm) as privacy becomes more and more stringent.

\subsection{Roadmap}
\color{black}
The remainder of the paper is organized as follows. Section~\ref{sec:background} provides background information and reviews related work in the area. Section~\ref{S:Stream} presents our proposed joint RF and camera-based CKD approach, and Section~\ref{sec:testbed} presents the technical setup of our experimental testbed. Section~\ref{sec:evaluation} presents our experimental evaluation details and provides insights from our findings. Section~\ref{sec:challenges} identifies key challenges and describes future research directions. Section~\ref{sec:conclusions} concludes this article.

%This project aims to understand the correlation of RF with video signals and explores the potential of using a combination of RF signals and machine learning technology to construct a test bench targeted at detecting patient movements/falls in a room. This project will subsequently apply our learnings for tuning and deploying SDR for accurate indoor localisation in the context of healthcare facilities. More specifically, we seek to understand and address the following research questions:

%\begin{itemize}
%    \item How can RF technology be leveraged to replace real-time camera utilisation for indoor localisation and patient monitoring?
%    \item What is the most appropriate machine learning model for understanding correlation between RF and video signals, and how can the model be suitably tuned?
%    \item How can SDR be appropriately and safely deployed to achieve indoor localisation and patient monitoring?
%\end{itemize}

%{Jonathan, perhaps this will help you. \color{red} Hinton et al.~\cite{hinton2015distilling} proposed knowledge distillation, which describes a learning style in which a more extensive Teacher NN guides a Student's NN learning process for numerous assignments. The understanding is passed on to Students by such Teachers using hidden labels~\cite{li2021online}.} 

\begin{figure}[t]
    \centering
    \includegraphics[width=3.9cm]{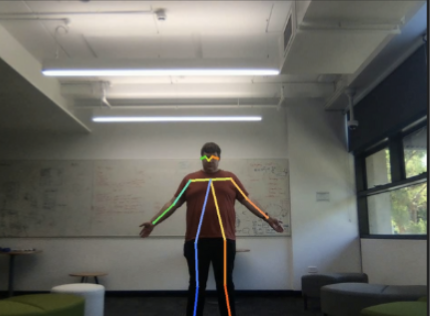}
     \includegraphics[width=4cm]{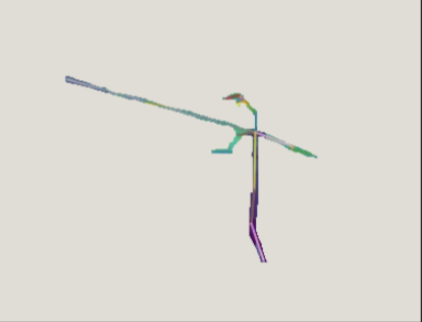}
    \caption{Observations without using proposed CKD framework. Left panel: Teacher class over original video frame with Alphapose detection. Right panel: Student class CSI trained model and pose detection using classic knowledge distillation}
    \label{fig:}
\end{figure}
\subsection{Literature}
\label{sec:background}
%In this section, we present the background on Software Defined Radio (SDR) and its application to the human activity sensing domain. We also present related works in the literature on using wireless signals for detecting human activities and movements, and the ML-based data analytical mechanisms that drive them.

In this section, we present the background information and related works pertaining to our findings presented in this paper.

\subsubsection*{Software-Defined Radio (SDR)} SDR has recently gained attention in the sensing and communication community due to its unique ability to unify two fundamental technologies, i.e., software and digital radio~\cite{tuttlebee_2002}. The work in~\cite{khan2022non} shows SDR as a radio communication system enabling the modeling and control of complex RF sensing tasks using a localizing environment to tune the waveforms. There are currently limited research outcomes documented in the literature on using SDR to monitor and identify human activities and/or pose changes.

%Patient monitoring might not only need fall detection, but we might also require to recognize various other activities and the of the most optimal ways in which we can do that is through human pose estimation. Human pose estimation has become quite popular in recent years. It can be performed using various sensing devices but what we must focus on are two such devices camera and Software-defined radio devices. 

%Can Radio Frequency technology (RF) be leveraged to replace real-time camera utilization for indoor localization and patient monitoring? This paper aims to understand the correlation of RF with video signals and explores the potential of using a combination of RF signals and machine learning technology to construct a test bench targeted at detecting falls in a room.

\begin{figure*}
    \centering
    \includegraphics[scale=0.34]{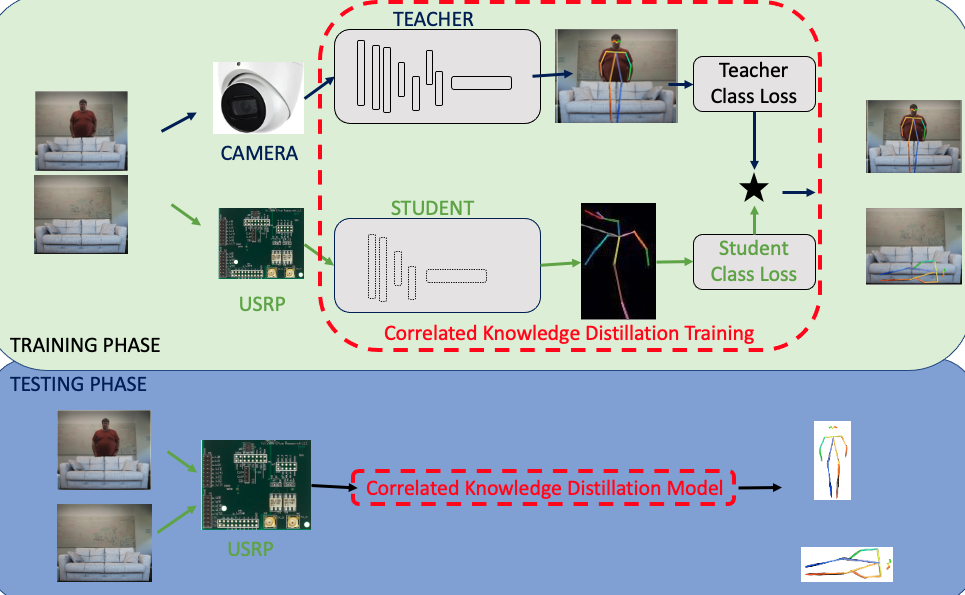}
    \caption{\color{black} An abstract view of the proposed correlated knowledge distillation framework. Observe that the classification prediction problem of the standard KD is separated into two levels: (i) a binary prediction for the camera-based Teacher class and all RF- sensing based Student classes and (ii) a multi-category prediction for each Student class. More importantly, we decompose the KD loss into two separate loss components.}
    \label{fig:CKD1}
\end{figure*}

\subsubsection*{Camera-based motion sensing}
The visual camera produces RGB images, and any human pose estimation from these images generally falls under top-down and bottom-up approaches. In a top-down approach, the camera first detects the person from the image, then applies a single-person pose estimator to each person to extract key points~\cite{fang2017rmpe,papandreou2017towards}. In a bottom-up approach, the camera first detects all the key points in the image first, then does some post-processing to associate the key points~\cite{pishchulin2016deepcut,cao2017realtime}. There are several approaches to using multi-person pose estimation from RGB images, for example, Alphapose~\cite{fang2017rmpe}, %CPN~\cite{chen2018cascaded}, Faster R-CNN~\cite{he2017mask}, SSD~\cite{liu2016ssd},
 Yolo~\cite{redmon2016you} and FPN~\cite{lin2017feature}. We exploit Aphapose (which uses a two-stage scheme) as our primary pose supervision resource which can ``teach'' the mapping from WiFi signals to a person's pose.

%This two-stage schema performs better than earlier methods based on global joint heat maps like OpenPose \cite{cao2017realtime}. Among all these approaches we utilize alpha pose as our main pose supervision to learn the mapping from WiFi signals to person pose.

\subsubsection*{WiFi-based human activity recognition}
There has been much interest in exploiting wireless signals for locating subjects and monitoring human activities. They can be classified into two categories. The first category utilizes very high frequencies e.g. millimeter wave (mmWave) or terahertz (THz)~\cite{zhu2015reusing}. It can realistically image the surface of the human body (such as airport security scanners), but it cannot penetrate objects such as walls and furniture. The second category utilizes lower frequencies around a few GHz, thus allowing it to scan through walls and occlusions. %Ground-breaking work in this field has explored and enhanced the capabilities of ``seeing through walls'' using WiFi~\cite{10.1145/2486001.2486039}. 

%Dina Katabi et al.~\cite{10.1145/2486001.2486039} propose using a human body's motion as an antenna array to track human motion by tracking the resulting RF beam.

{\color{black}
Important related works include crowd counting \cite{depatla2018crowd,hanif2018wispy}, human identification through physical constraints~\cite{cheng2019walls}, through-the-wall human detection~\cite{sun2021through,zhao2019through} %,di2018wifi},
or through-the-wall activity human activity or gesture recognition~\cite{yue2020bodycompass,wang2019joint,abdelnasser2018ubiquitous}. In particular, the work in~\cite{zhao2018through} proposes a state-of-the-art vision model to provide cross-modal supervision. The authors extracted pose information from the visual stream input and use it to guide the training process of the network model. Once the model is sufficiently trained, it will only use wireless signals for pose estimation. Another work in~\cite{wang2019joint} proposed a fully convolutional network to estimate multi-person pose from the collected data and analytics by training with the help of a visual stream from a camera.

The application use cases of the proposed mechanisms and techniques in this field are broad. The ability to accurately understand, correlate and fuse data, before performing (near) real-time prediction is invaluable. The large body of work in this field has seen new and emerging use cases. For example, the notable `RF-Sleep' work presented in~\cite{zhang2019smars,zhao2017learning} learns to predict sleep stages from radio measurements without any attached sensors on subjects. A predictive model combining convolutional and recurrent neural networks are developed to extract sleep-specific subject features from RF signals and progressively capture the temporal progression of sleep. An important aspect of the proposed method is the modified adversarial training regime that discards individual-specific information, which helps preserve the privacy of the subjects monitored. A recent work in~\cite{he2022contactless} proposed an innovative contactless approach to monitor patients' blood oxygen remotely by analyzing the radio signals in the room where the patients are located. This is performed without any sensors or wearables attached to the patients. The patient's respiration is being monitored from the radio signals that are reflected on their bodies. This mechanism uses a novel neural network that infers a patient's oxygen level from their breathing signal, with a gated transformer model. Advancements in coupling RF sensing and machine learning techniques have found important applications in healthcare, smart transportation and logistics, onset detection of diseases, agriculture, wildlife tracking, where privacy preservation is paramount~\cite{vasisht2018duet,tian2018rf,hsu2019enabling,li2019making,li2023mage,vonehr2016software}. The use of CKD is limited in the literature, which forms the primary focus and novel contributions of our work presented in this paper.
}

\subsubsection*{Knowledge Distillation (KD)} CKD proposed in this paper is an enhancement to KD by exploiting a knowledge fusion that extends a ``Teacher-Student'' approach to transfer knowledge from data/instances derived and collected from separate sources and transfer both instance-level and also the information of correlation between instances~\cite{peng2019correlation,li2020local, zhao2022decoupled}. To further understand this, we have tested standard KD~\cite{hinton2015distilling} Teacher-Student by plugging Alphapose~\cite{fang2017rmpe} into our experimental testbed. Our observations are shown in Figure~\ref{fig:}. We find that the inaccuracy in the pose estimates that we can see in the right panel of Figure~\ref{fig:} is due to RF noise and lack of knowledge fusion. In a different context, the authors in ~\cite{chen2021distilling, peng2019correlation,li2020local, zhao2022decoupled} proposed knowledge distillation (and with fusion~\cite{li2023knowledge}), which describes a learning style in which a more extensive Teacher Neural Network (NN) guides a Student's NN learning process for numerous assignments. This understanding is passed on to Students by such Teachers using hidden labels~\cite{li2021online} and fused multiscale distillation~\cite{li2023knowledge}.

To the best of our knowledge, there are currently no existing works in the literature that study the use of CKD by employing knowledge fusion distillation theory to correlate RF and visual signals (using both SDR and visual camera) and build an integrated deep model to rapidly identify and predict human pose changes. Our work builds on existing literature in RF/WiFi sensing of human activities, but extends to using SDR for RF signal generation and using CKD models for data correlation.

\section{Correlated Knowledge Distillation Design}    \label{S:Stream}

In this section, we present the design and technical details of our proposed joint CKD-RF sensing framework and approach for monitoring human pose.

\subsection{Proposed Correlated Knowledge Distillation (CKD) Framework}

With relevant insights from cutting-edge KD works~\cite{chen2021distilling, peng2019correlation,li2020local, zhao2022decoupled, li2023knowledge}, in our proposed CKD approach, the classification prediction problem in the underlying KD is separated into two correlated levels: (i) a binary prediction for the camera-based Teacher class and all SDR-based Student classes (RF-Sensing) and (ii) a multicategory prediction for each Student class. As shown in Figure~\ref{fig:CKD1}, using ideas from~\cite{zhao2022decoupled}, we decompose the KD loss into two components.
 
i)\textbf{ Teacher class loss:} In camera-based Teacher class training, the prediction of the Teacher class, denoted by $\mathcal T$, is provided, but the training of each Student class is based only on information transfer through binary logit distillation. One plausible explanation is that the KD Teacher class conveys information on the difficulty of training models with human poses; that is, the information specifies how challenging it is to recognize each pose. %We create experiments to test this hypothesis under three conditions meant to raise the "difficulty" of training data: more robust augmentation, label noise, and a naturally difficult dataset.
 
ii)\textbf{ Student class loss}: In Student Class, we consider only the knowledge among Student class logits that solely uses the outcomes that are equivalent to or better than those of the conventional KD, demonstrating the critical role of knowledge included in Student logits, which may provide insightful hidden information.
 
We have the following new information to develop the proposed CKD approach. The potential for logit distillation is constrained for some reasons, leading to uncorrelated results.  Furthermore, straightforward early feature fusion with logits does not make a competent Teacher direct their own training. We develop a module for correlated feature fusing and distillation, a novel technique for self-distillation. Logits are at more of a semantic and contextual level than deep network features; therefore, intuitively, logit distillation can achieve better efficiency than feature distillation. Therefore, we propose decomposing the celebrated KD mechanism towards correlated KD for multimodal learning. Such redesign of the KD approach enables us to investigate the effects of each component separately.
\begin{algorithm}[t]
\caption{\color{black} Details of the Pseudocode of CKD Algorithm}
\label{algo:CKD2}
%\begin{algorithmic}
\begin{lstlisting}[language=Python, numbers=none]
# temp: the temperature for KD and CKD
# a, b: hyper-parameters
.....................................
Creating class  CKD(Distiller)
        self.lossweight = cfg.DKD.CE-WEIGHT
        self.a = cfg.CKD.A
        self.b = cfg.CKD.B
        self.tem = cfg.CKD.T
        
Procedure Forward Train
        logitsStudent= self.Student(csi)
        logitsTeacher = self.Teacher(frame)
End Procedure
.....................................
Procedure LOSS
 lossce = lossweight*CrossEntropy(logits, target)
 lossCKD(epoch/warmup,CKDloss(logits,target,self)
EndProcedure

CKDloss(logitsStudent,logitsTeacher,target,a,b,tem)
    Teacher = F.softmax(logitsTeacher/tem)
    Student = F.softmax(logitsStudent/tem)
    Compute torch.log of (Teacher, Student) 
   TKDloss =F.NSELoss(Student,Teacher)

    Teacher1 = F.log-softmax(Teacher, mask(logits, target))
    Student1 = F.log-softmax(Student, mask(logits, target))
    SKDloss =F.NSELoss(Student1,Teacher1)
return (TKDloss + b * SKDloss)
.........................................
cat-mask(t): compute mask1, mask2
    rt = torch.cat([., .])
return rt
\end{lstlisting}
\end{algorithm}
  \subsection{Details of CKD}
  \label{sec:ckdtheory}
The details of the proposed CKD approach are shown as pseudocode in Algorithm~\ref{algo:CKD2}. For tractability~\cite{zhao2022decoupled}, we use the following notations to quantify the correlated features classification relevant and irrelevant to
the Student class, $\mathcal S$. The classification probability of the logit sample, with $l_i$ representing the logits of the $i$-th classification, relevant to the Student class, is given by $\mathbf{p}=[p_1, p_2,\dots, p_i, \dots, p_N]\in \mathbb{R}^{1\times N}$, where $N$ is the number of classifications and $p_i$ is the probability of the $i$-th classification obtained using \textit{softmax function}
\be
p_i= \frac{\exp(l_i)}{\sum^N_i \exp(l_i)}.
\ee
Consider $\mathbf{\bar p}=[p_r,p_{\bar r}]\in \mathbb{R}^{1\times 2}$ as the binary probabilities of the
relevant classification ($p_r$) and non-relevant classification ($p_{\bar r}$), which can be computed as
\begin{eqnarray}
    p_r&=& \frac{\exp(l_r)}{\sum^N_i \exp(l_i)};\\
p_{\bar r}&=&\frac{\sum^N_{i\neq r}\exp(l_i)}{\sum^N_i \exp(l_i)}
\end{eqnarray}
 In addition, without considering the relevant $r$-th classification, we quantify the probability $\hat p_i$ as
 \be
\hat p_i= \frac{\exp(l_i)}{\sum^N_{i\neq r} \exp(l_i)} 
\ee
for independently determining the model probabilities \[\mathbf{\hat p}=[\hat p_1, \hat p_2,\dots, p_{r-1}, p_{r+1} \dots, p_N]\in \mathbb{R}^{1\times (N-1)}.\]

In general, the novelty of the proposed CKD can be seen in the form of a tractable classification loss process as follows. To enable the Student model to learn the logit of the Teacher model directly, we compute the loss by applying a Kullback–Leibler (KL) divergence-like measure or the normalized squared error (NSE) between the logit vectors as
\be
NSE(\mathbf{\bar p_1},\mathbf{\bar p_2})=\Delta(\mathbf{\bar p_1},\mathbf{\bar p_2})=\frac{||\log\mathbf{\bar p_1}-\log\mathbf{\bar p_2}||^2}{N}.
\label{eqn:nse}
\ee
Observe in~\eqref{eqn:nse} that the normalization of the squared error of the CKD estimator is the average of the squared differences between the logarithmic estimated classification probabilities of the Teacher and Student class. It is a risk function representing the predicted cost of squared deviation, which is never negative, and a value closer to 0 is preferable. Considering that the 
%normalized squared error 
NSE is the second moment of the difference, 
%$MSELoss(\mathbf{\hat p_1},\mathbf{\hat p_2})=\sum(\mathbf{\hat p_1}-\mathbf{\hat p_2})^2/N$, 
$\Delta(\mathbf{\hat p_1},\mathbf{\hat p_2})=\frac{||\log\mathbf{\hat p_1}-\log\mathbf{\hat p_2}||^2}{N}$, we 
include both the estimator's variance and its bias, to reflect the importance of correlated multimodal learning in the CKD framework.

Using a similar intuition to that of KL-divergence (in the standard KD) for tractability, and using the aforementioned analyses~\cite{zhao2022decoupled}, we determine the CKD loss (a joint NSE loss of two classes) as
\begin{eqnarray}
CKDloss &=& TKDloss+(1-p_r^{\mathcal T})SKDloss,\\
   &=& NSE(\mathbf{\bar p}^{\mathcal S}, \mathbf{\bar p}^{\mathcal T})+(1-p_r^{\mathcal T})NSE(\mathbf{\hat p}^{\mathcal S}, \mathbf{\hat p}^{\mathcal T}). \nonumber
\end{eqnarray}
Intuitively,
$TKDloss$ is the feature classification loss,
$SKDloss$ is the model probabilities loss, and $p_r^{\mathcal T}$ is the relevant binary classification probability of the Teacher class network. To this end, the standard KD~\cite{hinton2015distilling} with knowledge fusion allows us to look at the distinct effects of TKD and SKD, highlighting the inadequacy of existing approaches~\cite{zhao2022decoupled} in exploiting the underlying correlation and multimodal feature fusion.

\subsection{Technical setup and design details}

\begin{figure}[h]
    \centering
     \includegraphics[width=2.2cm]{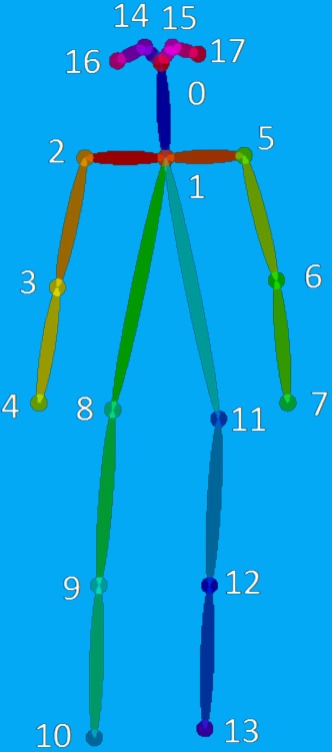}
   \includegraphics[width=4cm]{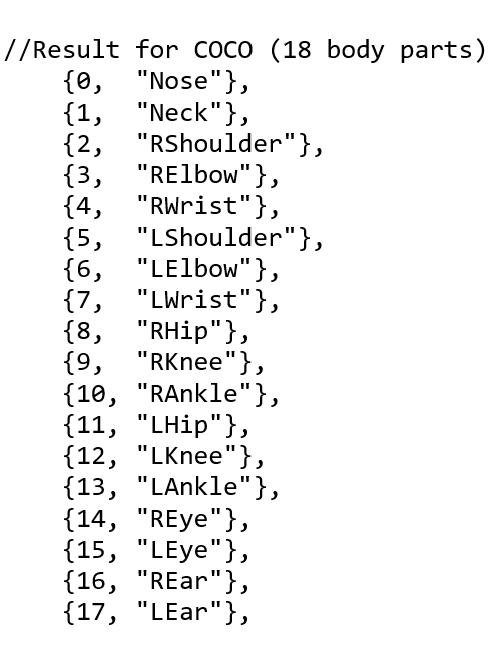}
    \caption{18 pose keypoint coordinates.}
   % \nocaption
    \label{fig: 18pose}
\end{figure}

To implement the proposed CKD, we extend AlphaPose~\cite{fang2017rmpe} to process the video and annotate the poses with confidence levels and coordinates for each frame accurately. This extension produces a JSON output file that contains person pose coordinates. The number of posing coordinates may differ depending on the dataset and model; nevertheless, we simply employ FastPose with the Yolov3 detector and Resnet50 as the backbone for tractability. Unlike the AlphaPose~\cite{fang2017rmpe} generating Teacher class model, we extract pose annotations and perform pose estimation from the WiFi Channel State Information (CSI) with learning to correlate with the Student class model. However, we find that we cannot accurately determine the pose coordinates of a person simply by analyzing the CSI. See Figure~\ref{fig:CKD1} for details. To ameliorate the shortcoming of such Student class training, we have implemented the proposed CKD approach, shown in Algorithm~\ref{algo:CKD2}, with a camera-oriented Teacher model training alongside the WiFi USRP to train the correlated model with videos and CSI.\footnote{The RGB video is captured using a laptop webcam at 20 frames per second and a 720p resolution, which saves the timestamps for each frame and synchronizes with the corresponding CSI.} For tractability~\cite{fang2017rmpe}, the underlying Alphapose of the Teacher model in our configuration returns 18 key points as shown in Figure~\ref{fig: 18pose} or pose coordinates and contributes in determining the corresponding Teacher Class Loss.
It is worth noting that these 18 key points are computed in the form of a JSON file format with a set of $\{x_1,y_1,c_1,...,x_k, y_k, c_k\}$.

% \begin{figure}[!h]
%     \centering
%    \includegraphics[width=6cm]{jsonoutputformat.png}.
%     \caption{Alphapose JSON File Format}
%     \label{fig: jsonoutputformat}
% \end{figure}
%\nointend is used for each person detected in the frame, and there will be another ``pose\_keypoints\_2d" field 

% \begin{enumerate}
%     \item Nose,
%     \item Neck,
%     \item RShoulder,
%     \item RElbow,
%     \item RWrist,
%     \item LShoulder,
%     \item LElbow,
%     \item LWrist,
%     \item RHip,
%     \item RKnee,
%     \item RAnkle,
%     \item LHip,
%     \item LKnee,
%     \item LAnkle,
%     \item REye,
%     \item LEye,
%     \item REar,
%     \item LEar,
%  \end{enumerate}

\color{black}
Our setup and design fully correlate the camera video with WiFi CSI which demonstrates the need for the proposed CKD approach. The WiFi CSI recording system consists of two ends, a sender with three antennas and a USRP with three antennas. When receiving WiFi transmissions, the USRP parses CSI~\cite{halperin2011tool} from WiFi signals. The parsed CSI is a tensor of size of ($\rm n \times 53 \times 3 \times 3$), where $n$ is the total number of signals received, $53$ is the subcarrier and the two $3's$ are the number of sender and receiver antennas, respectively. AlphaPose produces n three-element predictions in the format of ($x_{i}, y_{i}; c_{i}$), where n is the number of critical points to be estimated, $x_i$ and $y_i$ are the coordinates of the i-th keypoint, and $c_i$ is the confidence of the aforementioned coordinates. In experiments, we set $n =18$, and use the keypoint configuration as shown in Figure~\ref{fig: 18pose}.

\begin{figure}[t]
\centering

\includegraphics[scale=0.265]{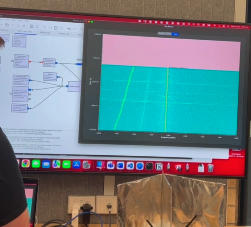}
\includegraphics[scale=0.212]{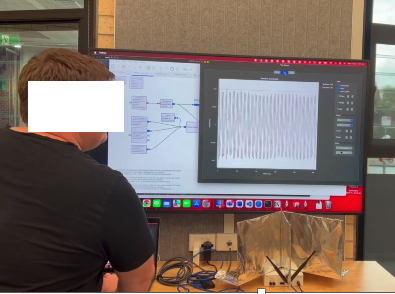}
\includegraphics[scale=0.2]{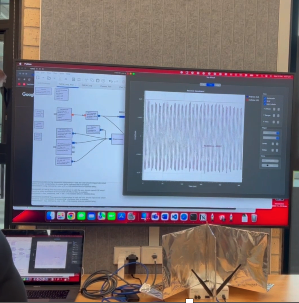}
 
 \includegraphics[scale=0.614]{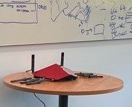}
 \includegraphics[scale=0.251]{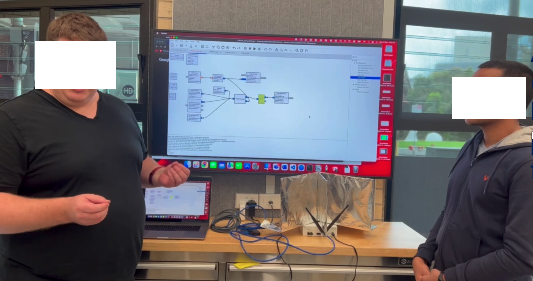}
 \includegraphics[scale=0.234]{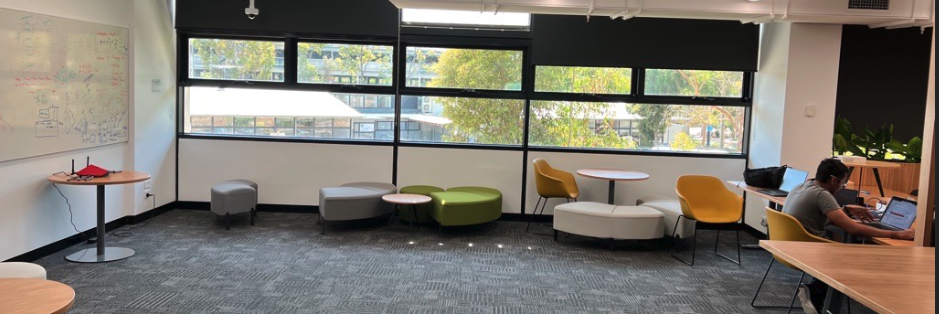}
\caption{Images of the testbed setup, videos and RF data collection. The images in the upper panel demonstrate our preliminary approach. The two bottom panels demonstrate our extended testbed setup and data collection. }
\label{expgrid}
\end{figure}

Assuming that $I_t$ and $C_t$ are a pair of synchronized image frames and RF CSI series, respectively, $t$ denotes the sampling time, the training dataset we generate is denoted by $D = (I_t, C_t), t \in [1, n].$ By developing CKD as described in Algorithm~\ref{algo:CKD2}, we trained D and learned their correlation. This is the learning mapping rules from the CSI series to the frame pose points. The Student class network $\mathcal S(.)$ fed with CSI, and the Teacher class network fed with AlphaPose, $\mathcal T(.)$, is the entire CKD framework. $\mathcal T(.)$ accepts $I_t$ as input, and using the person detector and posture regressor, returns the body keypoint coordinates and confidence, $(x_t, y_t; c_t)$, for each pair of $(I_t, C_t)$. The outputs are converted into an adjacent matrix to the body pose, or $PAM_t$. 1. Using $\mathcal T(I_t) \xrightarrow{} PAM_t$, we model how the Teacher network teaches $\mathcal S(.)$, where $PAM_t$ stands for cross-modality supervision to teach $\mathcal S(.)$.

%We go into great detail on S(.), i.e., WiSPPN. S(.) uses Ct as input during the training phase and produces a corresponding prediction of the adjacent posture matrix. S(.) is then optimized under $PAM_t$ supervision. WiSPPN comprises three essential modules: the encoder, feature extractor, and decoder. A $C_t$ is transformed into a $PAM_t$ prediction sequentially using these three modules. We will now go over the three modules' specific parameter information and our design goals.

In addition to the CKD, we have an encoder, feature generator, and decoder.  Encoder upsamples $C_t$ to proper width and height, suitable for the mainstream convolutional backbone networks and ResNet. Our CSI samples are 30 x 3 x 3 and the RGB frame is 3 x 244 x 244. Therefore, for CSI, by using eight gradually stacked transposed convolutional layers, we up-sampled $C_t$. The feature generator learns effective features to estimate the pose coordinates of a person using the upsampled $C_t$. To this end, we require a robust feature extractor to generate spatial information from the upsampled $C_t$, compared to an image frame, as CSI lacks spatial information of the key points of the body. We stack four basic ResNet blocks (16 convolutional layers) as feature generators to extract insights (a size of 300x18x18), which transforms the upsampled $C_t$ to $F_t$.

Our decoder is for the adaptation and correlation of the shape between the learned characteristic, $F_t$, and the supervised output of $\mathcal T(.)$, $PAM_t$. A body keypoint can be localized in the posture estimation job as two coordinates, namely the x and y axes. Therefore, the decoder is designed to take $F_t$ as input and predict the adjacency matrix within $(x,y)$ dimensions, resulting in a \textit{predicted pose adjacent matrix} $\overline {PAM_t}$. To accomplish this, we have two convolutional layers where the first layer is to release channel-wise information, and the second is to reorganize further the spatial information using kernels.

In fact, the Student class deep network predicts a pose adjacent matrix on each CSI input, $C_t$, using the encoder, feature extractor, and decoder. This process is designated as $S(C_t) \in \overline {PAM_t}$. Furthermore, with the CKD approach, the $PAM_t$ resulting from the Teacher network is feedback to supervise every predicted $\overline {PAM_t}$ during the Student training. %The Student network gains the ability to do single-person pose estimation only with CSI input when our training results in a satisfied performance of the Student network.  

\begin{figure}
    \centering
   \includegraphics[width=8cm]{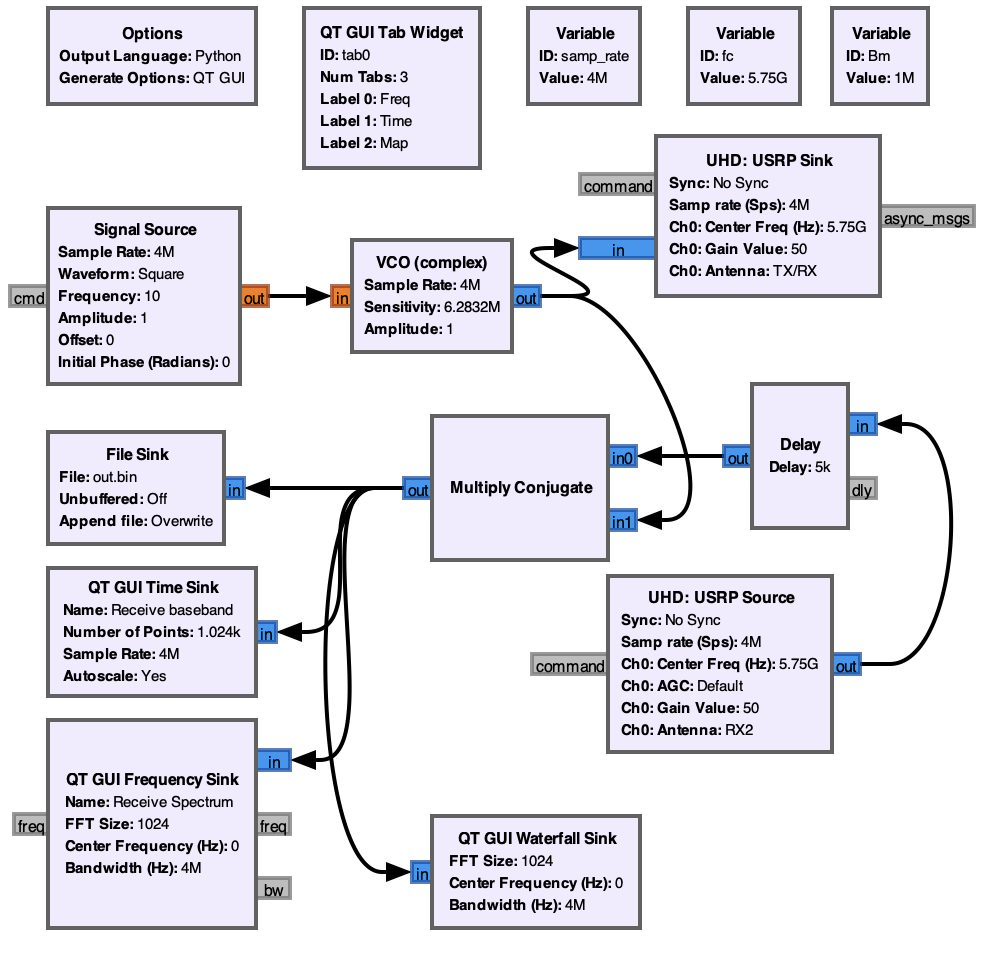}
    \caption{FMCW Radar design in GNU Radio.}
    \label{fig:GNU}
\end{figure}

% \subsection{Pose Adjacent Matrix Similarity Loss}
% The Teacher network outputs $PAM_t \in R^{3x18x18}$ as supervisions, and the Student network outputs $pPAM_t \in R^{2x18x18}$ as predictions from what we have seen previously. With the help of both supervision and predictions, we compute loss using the prediction confidence keypoints and L2 loss to optimise our Student network WiSPPN as shown below:
% \begin{equation}
%   \mathcal{L} = PAM^c * (||pPAM^x - PAM^x||_2^2 + ||pPAM^y - PAM^y||_2^2)
% \end{equation}
% where $||.||_2^2$ is the operator used to denote L2 distance. $pPAM^x$ and $PAM^x$ denote the prediction and supervision of pose adjacent matrix for body keypoint coordinate in the x-axis, respectively;$pPAM^y$ and $PAM^y$ are with similar representation while in the y-axis.

\section{Experimental Methodology}
\label{sec:testbed}

In this section, we present the setup of our SDR-based testbed and the implementation details of our proposed CKD framework.

Our testbed implements a USRP N200 with a UBX 40 USRP Daughterboard and two Omni-directional VERT2450 Antennas. Using this configuration, we first implemented an FMCW (Frequency-modulated continuous-wave) Radar utilizing the GNU radio, which is a free and open-source radio platform for generating and interpreting radio signals based on C++ and Python. Some of the images of our testbed setup and data collection are shown in the top panel of Figure~\ref{expgrid}. Figure~\ref{fig:GNU} shows our GNU radio configuration, which is extended with essential information from PiRadar. The SDR operates at 5.75 GHz and is fed with a saw tooth wave to modulate the carrier. We transmit the modulated carrier and record the reflected signal to detect state variations and predict pose estimates. This preliminary setup provides significant findings; however, we are unable to capture body movements from the binary data, which reflects very little of the pose changes. Also, the generated dataset is insufficient to feed deep neural network training and potential knowledge distillation.

With relevant insights from our first experiment, we extend the testbed to capture channel state information (CSI) using the 802.11ac WiFi standard. Images of our extended testbed setup and data collection are also shown in the two lower panels of Figure~\ref{expgrid}.  WiFi access points transmit orthogonally modulated beacons containing information about available WiFi, such as SSID, encryption type, and supported rates. Using the same SDR of our earlier mentioned setup, we first capture the beacons generated from a different access point in the environment to record the CSI values. We employ PicoScenes~\cite{9515545},\footnote{PicoScenes~\cite{9515545} is a middleware tool to capture the CSI and is compatible with our USRP-based testbed} to capture our CSI information. In this improved testbed, the USRP N200 is connected to a MacBook Pro-15-inch 2017 with Ubuntu 20.04 to facilitate PicoScenes capturing network traffic over frequency 5220MHz (5Ghz) and channel 44 driven by a Dlink DIR-895L access point. The video frames captured by the MacBook Pro are produced with a timestamp file for synchronization with the CSI. 

%Deakin has authorized all human subjects experiments, and 
We position the WiFi access points in our lab to have a clear line of sight from the USRP; 4 and 8 meters is the typical separation between the WiFi access points and the USRP. The project team has followed the Deakin University policies and processes for carrying out human subjects’ experimentation and data collection.  In the two Deakin Lab rooms on the Burwood campus, two authors (Satish and Williams) performed everyday pose tasks for five weeks. Figure~\ref{expgrid} shows floor plans and data collection locations. We use the setup in Figure~\ref{expgrid} to simultaneously record CSI samples and videos of 5-7 seconds while the pose actions occur (cf. Figure~\ref{fig:CKD1}). Each person's 85\% of videos and CSI recordings are used to train the networks, while 15\% of CSI are used to test the networks. Training and testing have data sizes of 72250 and 12750 videos and CSI samples, respectively. We have received written consent from all participants.% The implementation of the CKD framework and sample dataset  are publicly available on GitHub.\footnote{\url{https://github.com/Deakin-RF-Sensing}}

\subsection{Implementation challenges and training details}
All convolution and deep networks mentioned above are implemented with PyTorch 1.0. They are trained using the Adam optimizer for 20 epochs with an initial learning rate of 0.001. In the Teacher-Student training of the proposed CKD, we noted that the learning rate decreases by approximately 0.3 near the seventh, tenth, and 18th epochs. After training, we determine the diagonal elements in $\overline {PAM_t}$ as the body keypoint prediction, e.g., $x_k^* = \overline {PAM}_{1,k}$ and $ y_k^* = \overline {PAM}_{2,k} , k \in [1,18]$. %We use the following formulae to determine the diagonal elements in pPAM as the body keypoint prediction.

% For x-axis:
% \begin{equation}
%   x_k^* = pPAM_{1,k,k} , k \in [1,18]
% \end{equation}

% For y-axis:
% \begin{equation}
%   y_k^* = pPAM_{2,k,k} , k \in [1,18]
% \end{equation}

%\subsection{Coding Details}
\begin{algorithm}[t]
\caption{\color{black} Pseudocode of Residual Block developed for ameliorating the vanishing gradients impacts in the proposed CKD approach.}
\label{algo:VGP2}
\begin{lstlisting}[language=Python,numbers=none]
Creating class ResidualBlock(Neural Network Module)
    super(ResidualBlock,self)...
    self.bn1 = nn.BatchNorm2d(out_channels)
    self.relu = nn.ReLU(inplace=True)
    self.conv1 = conv3x3(in_channels,out_channels,.)
    self.conv2 = conv3x3(out_channels,out_channels)
    self.bn2 = nn.BatchNorm2d(out_channels)
forward(self, x)
    out = self.conv1(x)
    out = self.bn2(self.conv2(..)))
    out = self.relu(out)
return out
\end{lstlisting}
%\end{algorithmic}
\end{algorithm}

\subsubsection{Vanishing Gradient Problem}
We have found that the straightforward implementation and training of the proposed CKD approach suffer from vanishing gradient problems. Therefore, we develop a Residual Block module in PyTorch, which helps to overcome the vanishing gradient problem prevalent in deep neural network training. In general, the gradients keep decreasing as they are backpropagated through multiple layers leading to the vanishing problem, which creates difficulty for the optimizer to update the weights in the deeper network layers. The developed Residual Block module is used to establish a bypass connection that enables gradients to skip levels and directly propagate to deeper layers, thus alleviating the adverse impacts of such vanishing gradients.

The details of the developed Residual Block are shown in Algorithm~\ref{algo:VGP2}. It takes as input the number of input channels, the number of output channels, the stride, and an optional downsample layer for scaling the tensor dimensions. As outlined in Algorithm~\ref{algo:VGP2}, it has two convolutional layers with a batch normalization layer, a ReLU activation function, and two convolutional layers between each. The residual connection is either the input tensor or the output of the downsample layer.

  \begin{algorithm}[h]
\caption{\color{black} Pseudocode of the Extended ResNet Module for the CKD approach.}
\label{algo:VGP3}
\begin{lstlisting}[language=Python, numbers=none]
 Extending class ResNet(nn.Module):
        super(ResNet, self)
        self.conv = nn.Sequential(Conv2,BatchNorm,.)
        self.layer1 = self.make_layer(.)
        .........
        self.layer4 = self.make_layer(.)
        self.decode = nn.Sequential(.),
 forward(self, x):
        x = F.interpolate(.)
        x = self.layer1(x)
        .........
        x = self.decode(x)
 return x
\end{lstlisting}
%\end{algorithmic}
\end{algorithm}

\subsubsection{Revising ResNet for CKD}
We revise and extend the standard ResNet module along the lines of the developed Residual Block discussed above. The revision of the ResNet module is to take the developed Residual Block module as input with a list of the number of layers in each block. As shown in Algorithm~\ref{algo:VGP3}, our ResNet starts with a convolutional neural layer with 150 input and 150 output channels, followed by a batch normalization layer and ReLU activation. Adopting the \textit{make\textunderscore layer} function, it creates four residual blocks, each with increasing input and output channels. Observe in Algorithm~\ref{algo:VGP3} that the \textit{make\textunderscore layer} is fed with the residual block, the number of input channels, the number of blocks and the stride value, which produces a list of Residual Block modules.\footnote{ The first output block of the revised ResNet has the provided number of input channels and output channels, and succeeding blocks have output channels equal to the previous block's input channels. }

\begin{algorithm}[h]
\caption{\color{black} Interaction of the components of CKD Approach: How Algorithm 1 works with Algorithms 2 and 3}
\label{algo:VGP5}
\begin{lstlisting}[language=Python, numbers=none]
from models.CKD_resnet 
import ResNet, ResidualBlock, Bottleneck
    CKD = ResNet(ResidualBlock, [2, 2, 2, 2])
    resnet = ResNet(ResidualBlock, [3, 4, 6, 3])
    resnet = ResNet(Bottleneck, [3, 4, 6, 3])
    CKD = CKD.cuda()
criterion_L2 = nn.NSELoss().cuda()
optimizer = Optimize(Adam,CKDparam,learningrate)
scheduler = Optimize(MultiStepLR(optimizer),gamma)
    CKD.train()
\end{lstlisting}
\end{algorithm}

The decoding layer at the end of the developed ResNet has a convolutional layer with 300 input channels and 64 output channels. It is followed by a batch normalization layer and a ReLU activation function. An input tensor is upsampled using the \textit{F.interpolate} of the ResNet while the output tensor is created by first passing the tensor through the decoding layer.

\subsubsection{Interaction of Algorithm 1 with Algorithms 2 and 3}
The interactions of CKD with Algorithms~\ref{algo:VGP2} and \ref{algo:VGP3} are shown in Algorithm~\ref{algo:VGP5}. In our implementation, all other variables, parameters, and constraints such as \textit{Normalized Squared Error (NSE)}, Adam Optimizer, etc. are inherited for the standard ResNet class constructed with a total of 8 Residual Blocks.

% The num\textunderscore epochs variable is set to 10, and the learning\textunderscore rate variable is set to 0.01. The loss function used for optimization is Mean Squared Error (MSE) implemented using the `nn.MSELoss` class.

% The Adam optimizer takes as input the parameters of the model and the learning rate.

% A learning rate scheduler is initialized using the `torch.optim.lr\textunderscore scheduler.MultiStepLR` class, with milestones at epochs 7, 10, 15, 20, 25, and 30, and a gamma value of 0.5. This means that the learning rate will be multiplied by 0.5 at each milestone epoch.

% The `train` method is sets the model to training mode.

%heat map subsection
\section{Performance Evaluation and Analysis}
\label{sec:evaluation}
%\subsection{Data Collection}

We have implemented the CKD framework and used the dataset collected from the testbed. % and made it publicly available on GitHub.\footnote{\url{https://github.com/Deakin-RF-Sensing}}
Small video sequences of 5-7 seconds are captured with the timestamps alongside PicoScenes, producing a CSI file containing all WiFi frames. Each frame has several features. We extracted the system time from the raw CSI data. These data are then converted/stored as a NumPy array and fed into the CKD along with the matching video frames.
%\begin{figure}
%     \centering
%    \includegraphics[width=5cm]{Frequency Graph.png}
%     \caption
%     \label{fig:my_label}
% \end{figure}
% \begin{figure}
%     \centering
%    \includegraphics[width=5cm]{Signal Time Graph.png}
%     \caption{Signal/Time Graph}
%     \label{fig:my_label}
% \end{figure}
% \begin{figure}
%     \centering
%    \includegraphics[width=5cm]{Waterfall Spectrogram.png}
%     \caption{Waterfall Spectrogram}
%     \label{fig:my_label}
% \end{figure}
%\subsection{Results and Discussion}

% \textcolor{black}{Please include some representative results as graphs/plots here. You can also include the "sample image" mentioned.}

\begin{figure}[b]
\centering
\vspace{-5 mm}
 \includegraphics[scale=0.46]{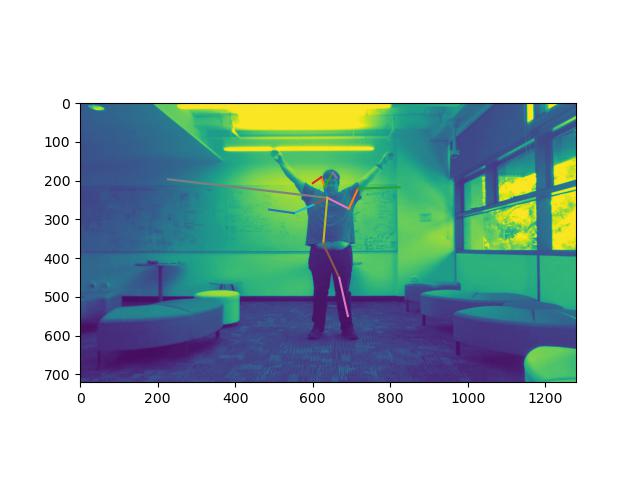}
\caption{Testing the Student class model using standard KD~\cite{li2021online} with test CSI data, pose estimates plotted  over the original frame. It is erroneous and demonstrates that only CSI with the KD model cannot quantify the poses accurately.}
\label{fig: dec}
\end{figure}

\begin{figure}[t]
\centering
 \includegraphics[width = 6cm]{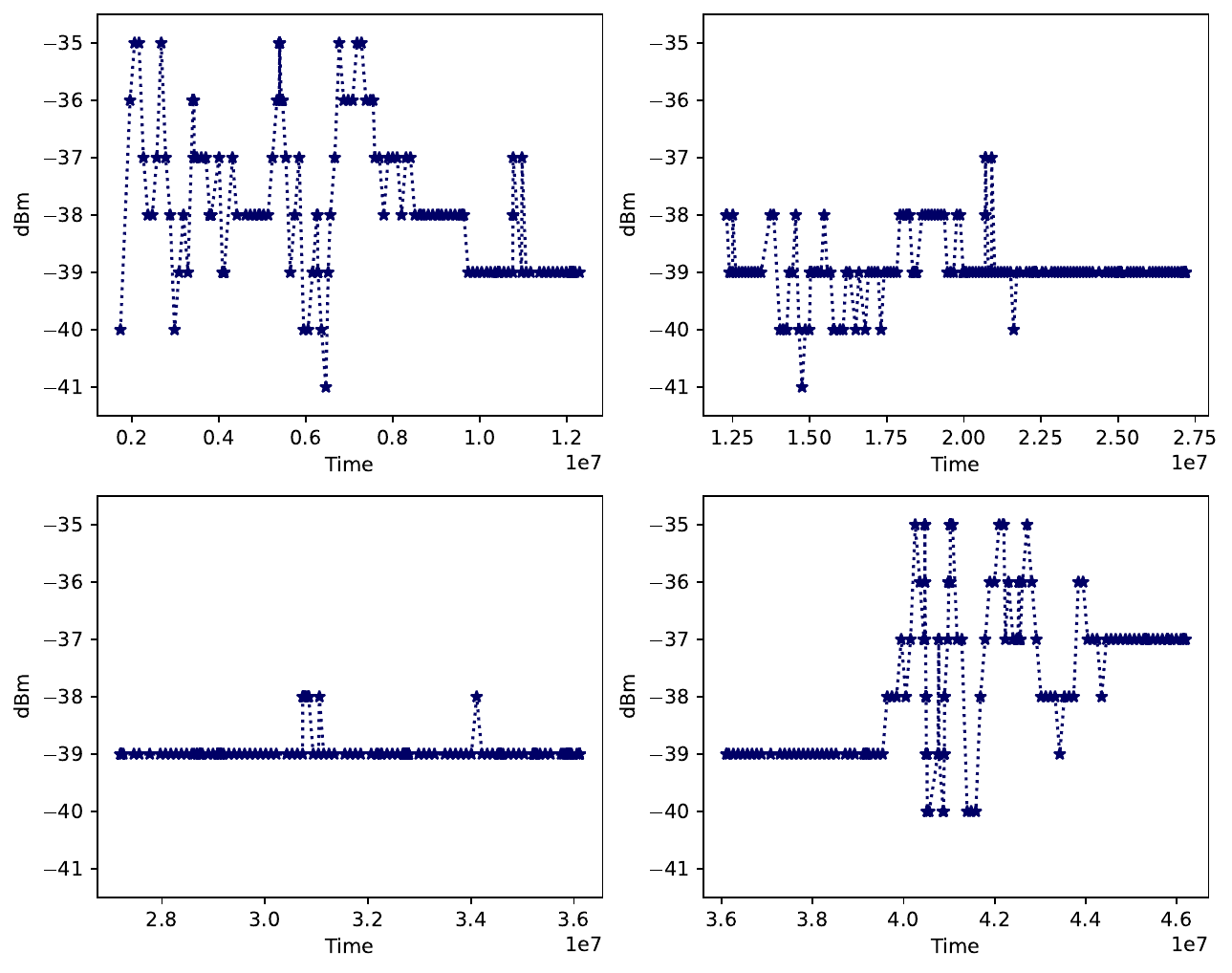}
\caption{\color{black} Signal Strength of CSI captured at USRP for the same pose in the same environment at different timestamps.}
\label{fig: rssi}
\end{figure}

\begin{figure}[t]
\centering
 \includegraphics[scale=0.34]{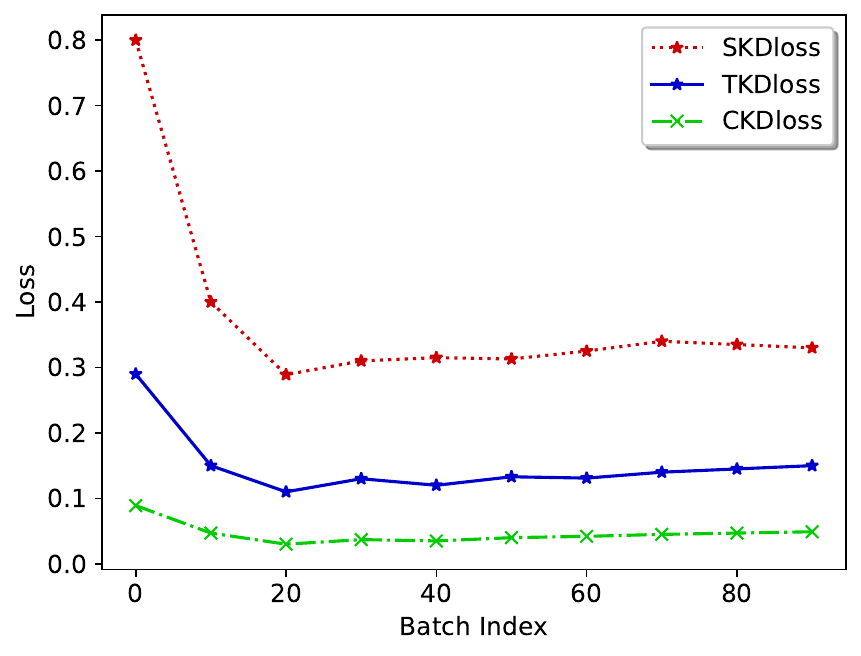}
\caption{\color{black} Evolution of the Teacher ($TKDloss$), Student ($SKDloss$) and CKD ($CKDloss$) loss over several epochs. The notable improvement (cf., $SKDloss \rightarrow CKDloss$) considerably increases the accuracy in pose detection, and most pose estimation testing obtained with the trained CKD model is much more accurate.}
\label{fig: lossgraph}
\end{figure}
\subsubsection{Benchmarking with standard KD} Using a standard KD~\cite{hinton2015distilling}, we performed the training and testing with the collected dataset for benchmarking. In Figure~\ref{fig: dec}, we can see that the pose estimation using standard KD is entirely inaccurate as the KD Loss framework is unable to capture the distinct difference in the classification loss of the video frame with that of CSI via the Student and the Teacher network model. In Figure~\ref{fig: rssi}, the corresponding variation in the signal strength of the CSI captured by the USRP at four different timestamps under the same posture of the same person is shown in Figure~\ref{fig: dec}. This is due to the adverse impact of RF noise and WiFi traffic fluctuations and demonstrates why the pose estimation directly from RF signals, such as using CSI shown in Figure~\ref{fig: dec}, is erroneous and is highly challenging (without knowledge fusion and correlated distillation from multimodal aspects).

\color{black}

\subsubsection{Training Efficiency of the Proposed CKD Framework}
\begin{table}[h]
    \centering
    \begin{tabular}{c|c|c|c}\hline
        \textbf{ Approach} &\textbf{Top-1 Acc}.& \textbf{Parameters} & \textbf{Time in ms}\\\hline
      KD~\cite{hinton2015distilling} &{70.23\%}& {Same parameters} & {10 milliseconds}\\
      R-KD~\cite{chen2021distilling} &{72.6\%}& {1.8 Million +} & {22 milliseconds}\\
       CKD~\cite{chen2021distilling} &{76.46\%}& {Same parameters} & {11 milliseconds}
    \end{tabular}
    \caption{\rm \color{black} Training time and accuracy of the CKD compared to others~\cite{hinton2015distilling, chen2021distilling} with the collected data set in our setup as explained in Sec.~\ref{sec:testbed}.}
    \label{tab1}
\end{table}

To demonstrate the excellent training efficiency of the proposed CKD, we compare training costs with other cutting-edge distillation techniques~\cite{hinton2015distilling, chen2021distilling}. Our CKD produces optimal results by optimizing the trade-off between model performance and training costs such as training time and additional parameters. The training time, top-1 accuracy, and additional parameter requirements of the proposed CKD framework with those of others~\cite{hinton2015distilling, chen2021distilling} are tabulated in Tab.~\ref{tab1}. Observe that, as a reformed version of the classic KD~\cite{hinton2015distilling}, CKD requires virtually the same computing complexity as KD and, indeed, the same set of parameters. Almost all celebrated feature-based distillation algorithms (\cite{chen2021distilling} and its enhancements) need additional training time to distill intermediate layer features, a huge number of additional parameters, and comparatively high GPU memory expenses, which are impractical in the context of the Deakin RF-Sensing studied in this work.

\begin{figure}[t]
\centering

\includegraphics[scale=0.3765]{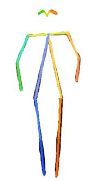}
\includegraphics[scale=0.3765]{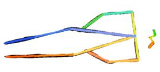}
\includegraphics[scale=0.3765]{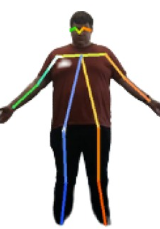}
 
 \includegraphics[scale=0.35]{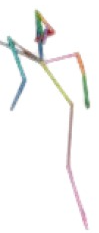}
 \includegraphics[scale=0.35]{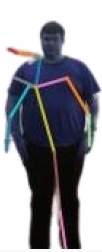}
 \includegraphics[scale=0.35]{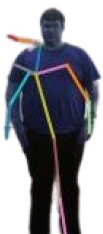}
 \includegraphics[scale=0.35]{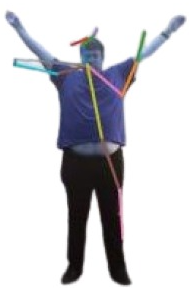}
\caption{Testing accuracy of the proposed CKD framework. Top panel: highly accurate pose estimates; bottom panel: relatively low accurate pose estimates; both using the test samples over the same experimental testbed.}
\label{testgrid}
\end{figure}

With the proposed CKD approach, we have successfully minimized the training loss considerably over the epoch index, as shown in Figure~\ref{fig: lossgraph}. We can see the differences in the evolution of the Teacher ($TKDloss$), Student ($SKDloss$) and CKD ($CKDloss$) loss over several epochs. This notable improvement (cf. $SKDloss \rightarrow CKDloss$) considerably increases the accuracy in pose detection, and most pose estimation tests obtained with the trained CKD model are much more accurate. We have shown some examples of our test images with the accuracy of the estimation of the CKD in Figure~\ref{testgrid}. %We observe that the low accuracy of in some of the test samples is due to the adverse impact of RF noise and WiFi traffic fluctuations.

\begin{figure}[t]
\centering
 \includegraphics[scale=0.35]{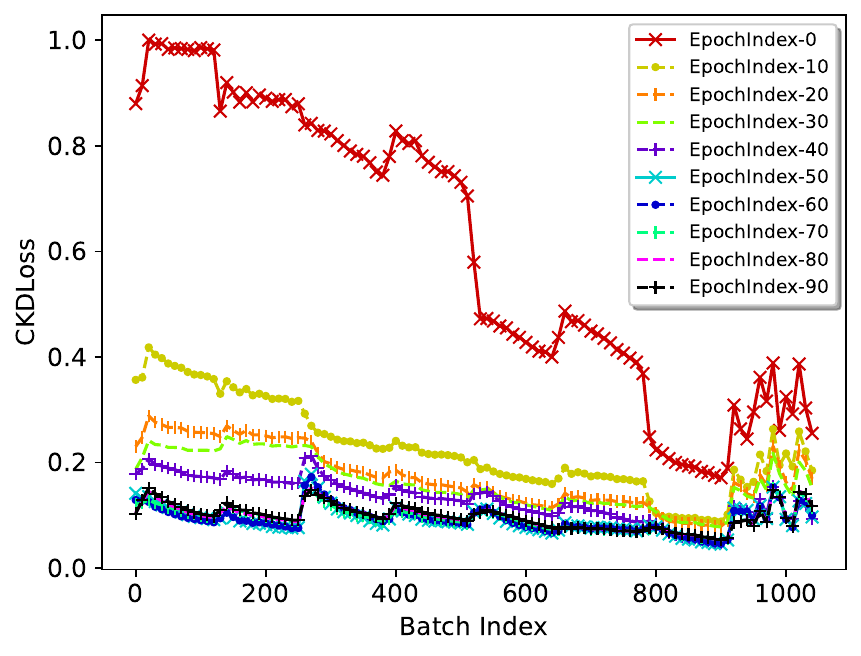}
\caption{Comparison of the $CKDloss$ over each batch index of data in different epochs as observed in our testing setup.}
\label{fig: decoder}
\end{figure}

To further understand this, we compute the MSE for each batch in different training epochs with the proposed CKD. As shown in Figure~\ref{fig: decoder}, we found that CKD can exploit correlation to minimize loss by considerably increasing the number of batches. Sample training and testing data can be found in the corresponding GitHub repositories. We have seen practically how the CKD method accelerates learning and improves accuracy. Our experiment shows that the CKD model successfully learns a person's posture when the person is relatively stationary and moving slowly but struggles when the person's posture is rapidly changing (as shown in Figure~\ref{testgrid}).

\color{black}
\textbf{Remark III.1 (Monitoring Rapidly changing Poses).} In our testbed, the RF data are captured differently (as complex numbers) and are noticed from different angles (horizontal and vertical projections) than the video frames, as the spatial \color{black}resolution of RF signals is substantially lower than the video signals (e.g. wall-penetrating frequencies). It is known that when the wavelength is longer than the surface roughness, a physics phenomenon known as RF-specularity takes place. Therefore, our body is specular in frequencies that may even go through walls. Intuitively, in the context of this research and our testbed setup, our body behaves more like a reflector (a mirror) than a scatterer. With a wavelength of around 3-5cm, humans serve as mirrors. The signal's reflection direction depends on whether or not each limb's surface is toward our USRP. It is worth noting that RF differs from a camera system in that a video frame only contains information on a subset of our limbs. We have observed that, because of a single USRP sensor in our testbed, when a person is running, we sometimes miss limbs and body parts whose orientation at that moment echoes the signal away from the USRP. We anticipate that tracking such rapidly changing aspects requires sophisticated testbed development and appears plausible only with multiple USRPs at spatially dispersed locations to fill such voids/misses. %Development of such a sophisticated testbed for monitoring rapidly changing pose dynamics using multiple USRPs and understanding their intra correlations with enhanced CKD framework requires further investigation in the future based on the findings in this work. This is beyond the budget and scope of the current project. 
%\color{black}

% \begin{figure}[H]
% \centering
%  \includegraphics[width = 8cm]{RSSI_4x4.png}
% \caption{RSSI (Signal Strength of CSI captured)}
% \label{fig: decoder}
% \end{figure}

% \begin{figure*}[t]
% \centering
%  \includegraphics[width=1\textwidth]{GridImage.jpg}
% \caption{Result Samples}
% \label{fig: decoder}
% \end{figure*}

\section{Applications, Challenges and Future Directions}
\label{sec:challenges}

{\color{black}
Initial findings and experimental insights of our proposed CKD-RF framework demonstrated the feasibility and practicality of correlating two (or more) independent data sources. Our proposed CKD-RF framework has broad application domains, spanning healthcare, smart transportation/logistics, mission-critical applications, education, and so forth. In addition to our motivating scenario of privacy-preserving patient monitoring, the use of our framework can be directly applicable to those similar to previous pioneering work in the field, including sleep monitoring~\cite{zhang2019smars,zhao2017learning}, onset detection of diseases~\cite{he2022contactless}, pose estimation~\cite{zhao2018rf} with fall detection use cases in homes / care / educational centers~\cite{vasisht2018duet,tian2018rf,hsu2019enabling,li2019making,li2023mage}. With advanced enhancements and improvements, our framework could be extended to other avenues such as vehicular traction vehicular traction control (with multiple sensors fed into the CKD model), mission-critical applications for subject localization, contact tracking, and movement control during disease outbreak, accurate positioning and localization unmanned aerial vehicles (UAVs) for goods delivery, wildlife tracking~\cite{vonehr2016software}, etc.
}

\color{black}
Although our findings improved with CKD, the CSI data we collect contain only one dimension because we only employ a single WiFi transmitter and USRP receiver. Therefore, the proposed CKD framework can only accurately extract crucial spatial information by reshaping and upsampling the poor one-dimensional CSI. Future work should expand this study to include scenarios with multiple access points and multiple USRP to test the feasibility of our proposed framework.

{\color{black}
There are also challenges surrounding the unwanted impacts of RF noise in our experiments. There are some well-known and documented alternative approaches for ameliorating the adverse impacts of RF noise that could be considered in our future experiments. For example, it is feasible to use Ultra-Wideband (UWB) technology~\cite{aiello2003ultra,sahinoglu2008ultra} to generate extremely narrow pulses or very short-duration signals which will mitigate the influence of radio interference and enable the signals to coexist with signals generated by other wireless systems. Our USRP SDR (and similar SDR and cognitive radio-based technology)~\cite{ulversoy2010software} allows us to adjust the transmitting/receiving frequencies, modulation schemes and other associated parameters. By incorporating the ability to dynamically adjust and adapt these factors and parameters make our system more resilient to RF noises. 
}

{\color{black}
The selection of frequency hopping and spread spectrum techniques (such as Direct Sequence Spread Spectrum and Frequency Hopping Spread Spectrum) within the generated RF signals in the SDR~\cite{rouphael2009rf,kohno2002signal} will intrinsically reduce the impact of RF noises. Our CKD-RF framework in conjunction with multiple directional antennas (such as those used in antenna arrays)~\cite{rahman2012fully,ashleibta2020non,goverdovsky2016modular} for data transmission/reception will significantly improve the accuracy by concentrating the signal towards a particular receiver, thus improving the signal-to-noise ratio. In addition, new SDR devices can enable dynamic spectrum access~\cite{rashid2011enabling,clendenen2012software} where the transmission frequency and power levels are intelligently adapted to the environment to avoid areas with high levels of noise and congestion. Unused frequencies can be quickly identified and opportunistically utilized.
}

We understand that the RF noise in the proposed sensing system restricts how stable the RF field can become. We have found that the RF controller can dampen noise, but still affects the cavity fields. The stability of the RF field is affected by feedback from the SDR detector noise, which in turn, impacts the accuracy and precision of the field measurements. A thorough investigation of the noise's origins and effects on the RF field is necessary to establish the RF system's performance requirements. For such a kind of quantification and development of tightly coupled approaches with CKD for noise amelioration, one needs to develop noise transfer relationships in RF control loops. In addition, noise models of several essential RF components (like amplifiers, mixers, and analog-digital converters) are to be developed, which indicates an important research direction to be considered in the future.

There is a great deal of potential in this area and in the ways it may be used. Still, we encountered some challenges (as discussed earlier) during implementation that are now resolved, and we can confidently use the proposed CKD approach. We have identified the following three important directions:
\begin{enumerate}
\item Our preliminary experiments and studies (along the lines of Figure~\ref{fig: rssi} and~\ref{testgrid}) with two access points suggest multiple access points  and USRP receivers can potentially generate higher accuracy and capture a spatial offset to quantify the dynamic change in CSI efficaciously, which requires extensive experiments in the future.
\item We have observed that the volume of WiFi traffic fluctuates over time and cannot generate enough RF signals for effective passive listening. Therefore, we need to look for an alternative approach that can generate more RF steadily with more success and ameliorate the adverse impacts of RF noise. Such an approach can extend the proposed CKD along the lines of the captured signals and by capturing seamless datasets. 
\item It appears possible to generate a precise RF signal and continuously monitor it from the access points. An additional module on the top of the CKD can help understand and predict the received RF signal, which can be more easily detected for deviations.
\end{enumerate}

\section{Conclusions}
\label{sec:conclusions}

%In this work, we have developed a Correlated Knowledge Distillation (CKD) system to teach inhouse wireless network of WiFi and USRP to sense human postures and movements. CKD system aims to realize a hybrid RF and camera framework for privacy-preserving human activity detection using software-defined radios. The CKD training comprised two parallel classification training with the images and radios for the aggregate knowledge fusion and distillation. Once trained, using radios from SDR, the CKD model can efficiently preserve privacy and utilize the multimodal correlated logits from the Teacher to the Student network without using the images and video frames. Through testing and validation using our SDR-based experimental testbed, we have demonstrated that the developed CKD framework is feasible and it's advancements have the potential for high impact in the field. 

In this work, we developed a Correlated Knowledge Distillation (CKD) system to teach our inhouse wireless network, comprising WiFi and USRP, to infer human postures and movements. The CKD system aimed to achieve a hybrid RF and camera framework for privacy-preserving human activity detection using only RF signals. CKD training involved two parallel classification training approaches with images and radios, allowing the fusion and distillation of knowledge. Once trained, the CKD model effectively preserved privacy and leveraged the correlated multimodal logits from the Teacher to the Student network, eliminating the need for images and video information. Through testing and validation using our SDR-based experimental testbed, we demonstrated the feasibility of the developed CKD framework, highlighting its potential for significant impact in the field.
Finally, we also note that the method of training using correlated twin neural networks, where one is a ``Teacher network'', is useful for leveraging pre-trained networks in the absence of (or  limited)  data.
\color{black}
%Software and dataset from this project at: \url{https://github.com/Deakin-RF-Sensing}.

%Although monitoring patients is the primary motivation for this work, our proposed CKD-RF system and framework have many potential applications outside of healthcare, including smart agriculture, transportation, and educational environments.

% \section*{Software and Dataset}
% \begin{flushleft}
% The software and dataset developed as part of this project can be found at:
% \hyperlink{https://github.com/MPTCP-FreeBSD/SDR-RF}{github.com/MPTCP-FreeBSD/SDR-RF}
% \end{flushleft}

%\section*{Acknowledgements}
%This work is partly supported  under the research program ``\textit{Next-generation IoT communications infrastructure, analytics and intelligence}'' at the School of IT of Deakin University. We appreciate the time and efforts of the two authors (Satish and Williams) who helped build our datasets and perform coding as human subjects. We appreciate the constructive criticism provided by the anonymous reviewers.

\bibliographystyle{ieeetr}
\bibliography{urllc}

\end{document}